\documentclass[letterpaper, 10 pt, conference]{ieeeconf}  % Comment this line out if you need a4paper

\IEEEoverridecommandlockouts                              % This command is only needed if 
\usepackage[hyphens]{url}
\usepackage{afterpage}
\usepackage[noadjust]{cite}
\usepackage{hyperref}  
\usepackage{amsmath}
\usepackage{times}
\usepackage{amssymb}
\usepackage{dsfont}
\usepackage{algpseudocode}
\usepackage{algorithm}
\usepackage{subcaption}
\usepackage{graphicx}
\usepackage{float}
\usepackage{booktabs}
\usepackage{ragged2e}
\usepackage{adjustbox}
\usepackage{bm}
\usepackage{multirow}
\usepackage{comment}
\usepackage{overpic}
\usepackage{soul}
\usepackage[dvipsnames]{xcolor}
\usepackage{wrapfig}

\title{\LARGE \bf
Manifold-constrained Hamilton-Jacobi Reachability Learning for Decentralized Multi-Agent Motion Planning}

\author{Qingyi Chen$^{1}$, Ruiqi Ni$^{1}$, Junyoung Kim$^{1}$ and Ahmed H. Qureshi$^{1}$ % <-this % stops a space
\thanks{*This material is based upon work supported by the Air Force Office of Scientific Research under award number FA9550-24-1-0233. Any opinions, findings, and conclusions or recommendations expressed in this material are those of the author(s) and do not necessarily reflect the views of the United States Air Force.}% <-this % stops a space
\thanks{$^{1}$Qingyi Chen, Ruiqi Ni, Junyoung Kim and Ahmed H. Qureshi are with the Department of Computer Science at Purdue University, West Lafayette, IN 47907, USA. {\tt\small \{chen5221, ni117, kim3722, ahqureshi\}@purdue.edu}}%
}

\begin{document}

\maketitle
\thispagestyle{empty}
\pagestyle{empty}

\newcommand{\qc}[1]{{\textnormal{\color{YellowOrange}\textbf{Qingyi: #1}}}}
\newtheorem{theorem}{Theorem}
\newtheorem{defn}[theorem]{Definition}
\newtheorem{assum}[theorem]{Assumption}
\newtheorem{rem}[theorem]{Remark}
\newtheorem{thm}[theorem]{Theorem}

\providecommand{\Opt}{\texttt{(Opt)}}
\providecommand{\optref}{\hyperref[opt]{\Opt{}}}
\providecommand{\methodname}{HaMMAR}

\begin{abstract}

Safe multi-agent motion planning (MAMP) under task-induced constraints is a critical challenge in robotics. Many real-world scenarios require robots to navigate dynamic environments while adhering to manifold constraints imposed by tasks. For example, service robots must carry cups upright while avoiding collisions with humans or other robots. Despite recent advances in decentralized MAMP for high-dimensional systems, incorporating manifold constraints remains difficult.
To address this, we propose a manifold-constrained Hamilton-Jacobi reachability (HJR) learning framework for decentralized MAMP. Our method solves HJR problems under manifold constraints to capture task-aware safety conditions, which are then integrated into a decentralized trajectory optimization planner. This enables robots to generate motion plans that are both safe and task-feasible without requiring assumptions about other agents’ policies.
Our approach generalizes across diverse manifold-constrained tasks and scales effectively to high-dimensional multi-agent manipulation problems. Experiments show that our method outperforms existing constrained motion planners and operates at speeds suitable for real-world applications. Video demonstrations and source code are available at \url{https://youtu.be/RYcEHMnPTH8} and \url{https://github.com/qingyichen/hammar}.

\end{abstract}

\section{Introduction}

Real-world operations often require multiple robots to function in a shared workspace while dynamically avoiding collisions and adhering to task-induced manifold constraints. For instance, teams of robots carrying large objects on a factory floor must maintain their closed-loop kinematic constraints while also avoiding collisions with other robots, whether those are moving independently or under similar constraints. Likewise, in restaurants, multiple service robots must maintain the orientation of trays or objects to prevent spillage while navigating to different tables without interfering with each other. The ability to plan and react in decentralized, multi-agent environments under manifold constraints is essential for realizing practical, large-scale applications of multi-robot systems.

Recent work on decentralized multi-agent motion planning (MAMP) has made promising progress in scaling to high-dimensional systems \cite{ha2020learningdecentralizedmultiarmmotion, Lai_2025, Gafur_2022, cooperative2020, Hartmann_2023, bakker2023multirobotlocalmotionplanning, rodriguez2024guaranteed}. However, these methods generally do not account for manifold constraints and assume the robots operate in the full configuration space, limiting their applications in  constrained settings. A complementary line of research focuses on developing centralized approaches to solving MAMP problems with manifold constraints. These methods typically extend sampling-based constrained motion planners \cite{Berenson-2009-10209, kingston2018sampling, jaillet2012asymptotically, berenson2011task, kim2016tangent} to multi-agent settings \cite{sampling-dual-arm, guo2024efficientmultirobotmotionplanning}, combining constrained sampling with centralized coordination or scheduling to generate feasible motions under task constraints. While effective in enforcing manifold constraints, their reliance on centralization limits their applicability in dynamic environments where agents move independently.

\begin{figure}[t!]
\centering
\begin{subfigure}[t]{0.16\textwidth}
\begin{overpic}[width=\textwidth]{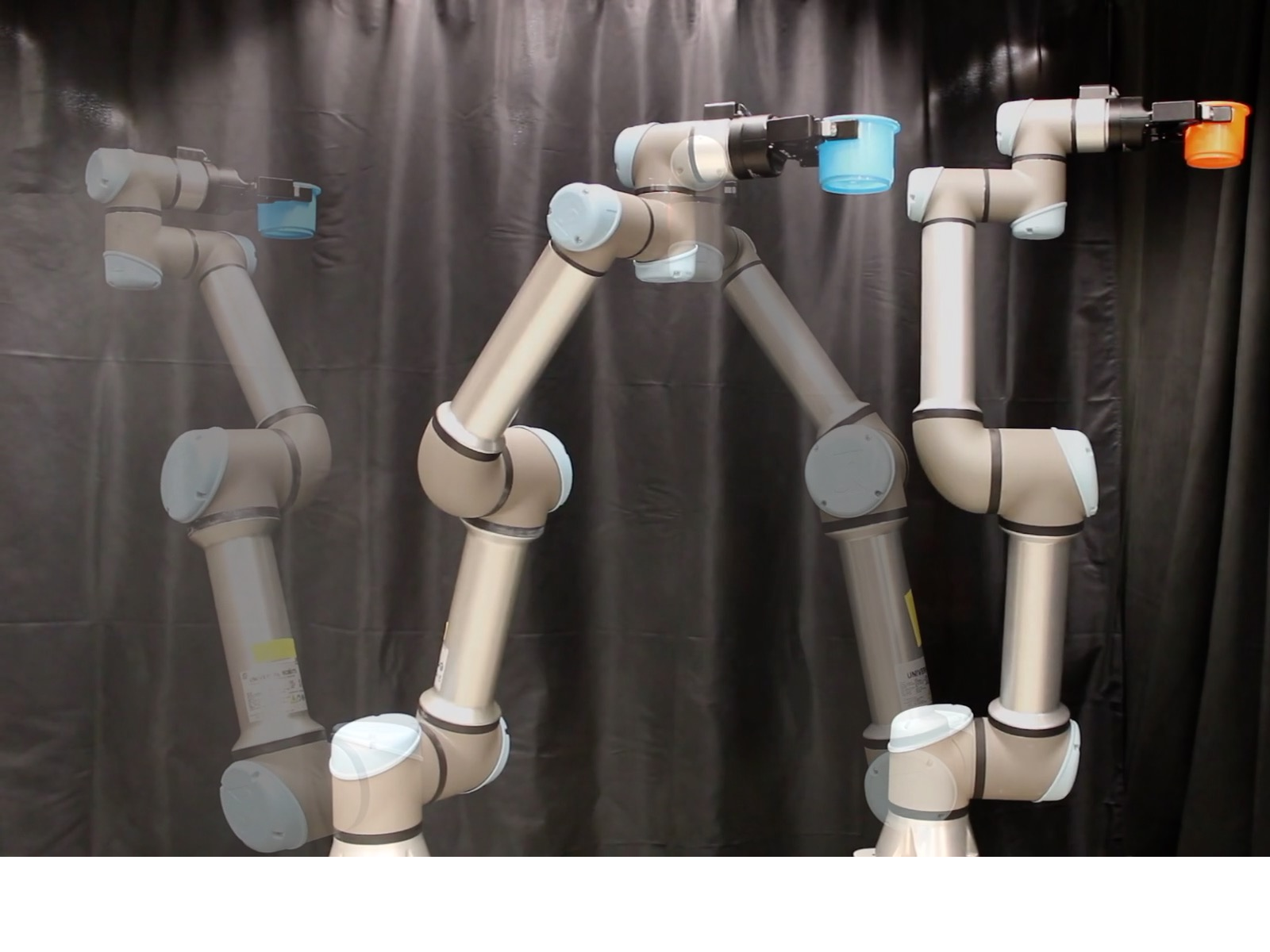}
\put(2,10){\color{white}\bfseries\small (1)}
\end{overpic}
\end{subfigure}\begin{subfigure}[t]{0.16\textwidth}
\begin{overpic}[width=\textwidth]{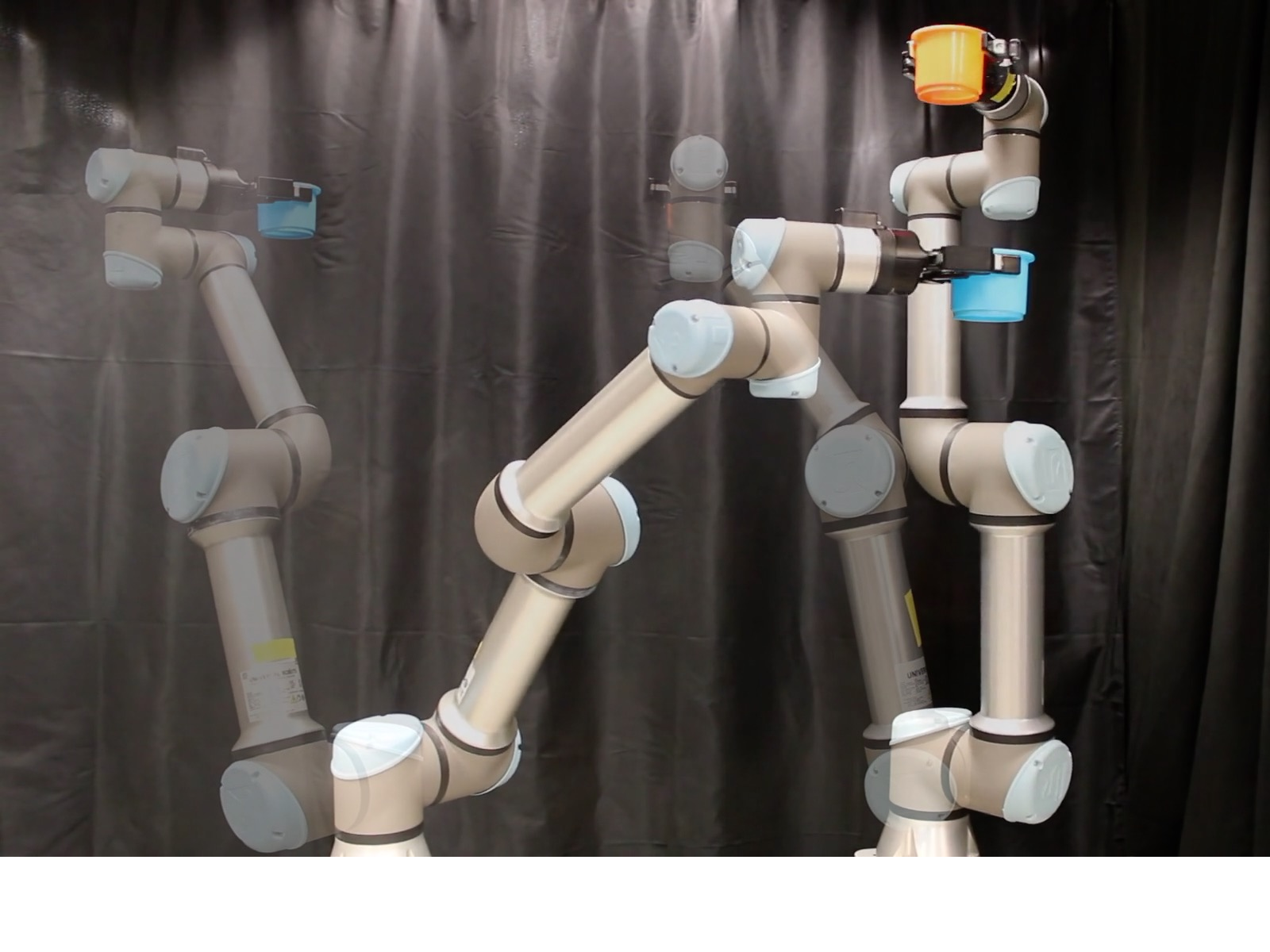}
\put(2,10){\color{white}\bfseries\small (2)}
\end{overpic}
\end{subfigure}\begin{subfigure}[t]{0.16\textwidth}
\begin{overpic}[width=\textwidth]{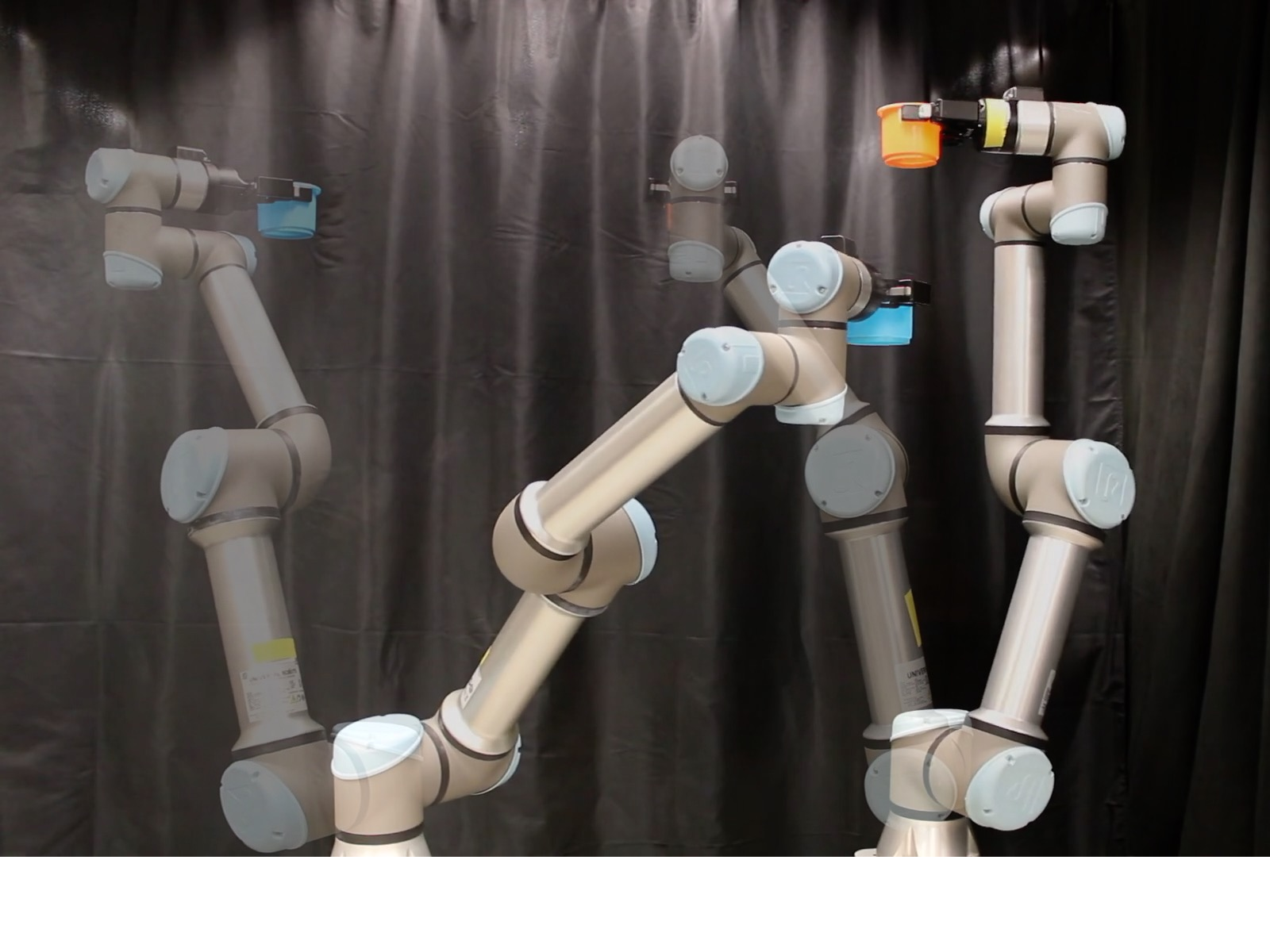}
\put(2,10){\color{white}\bfseries\small (3)}
\end{overpic}
\end{subfigure}
\vskip -1.2ex
\begin{subfigure}[t]{0.16\textwidth}
\begin{overpic}[width=\textwidth]{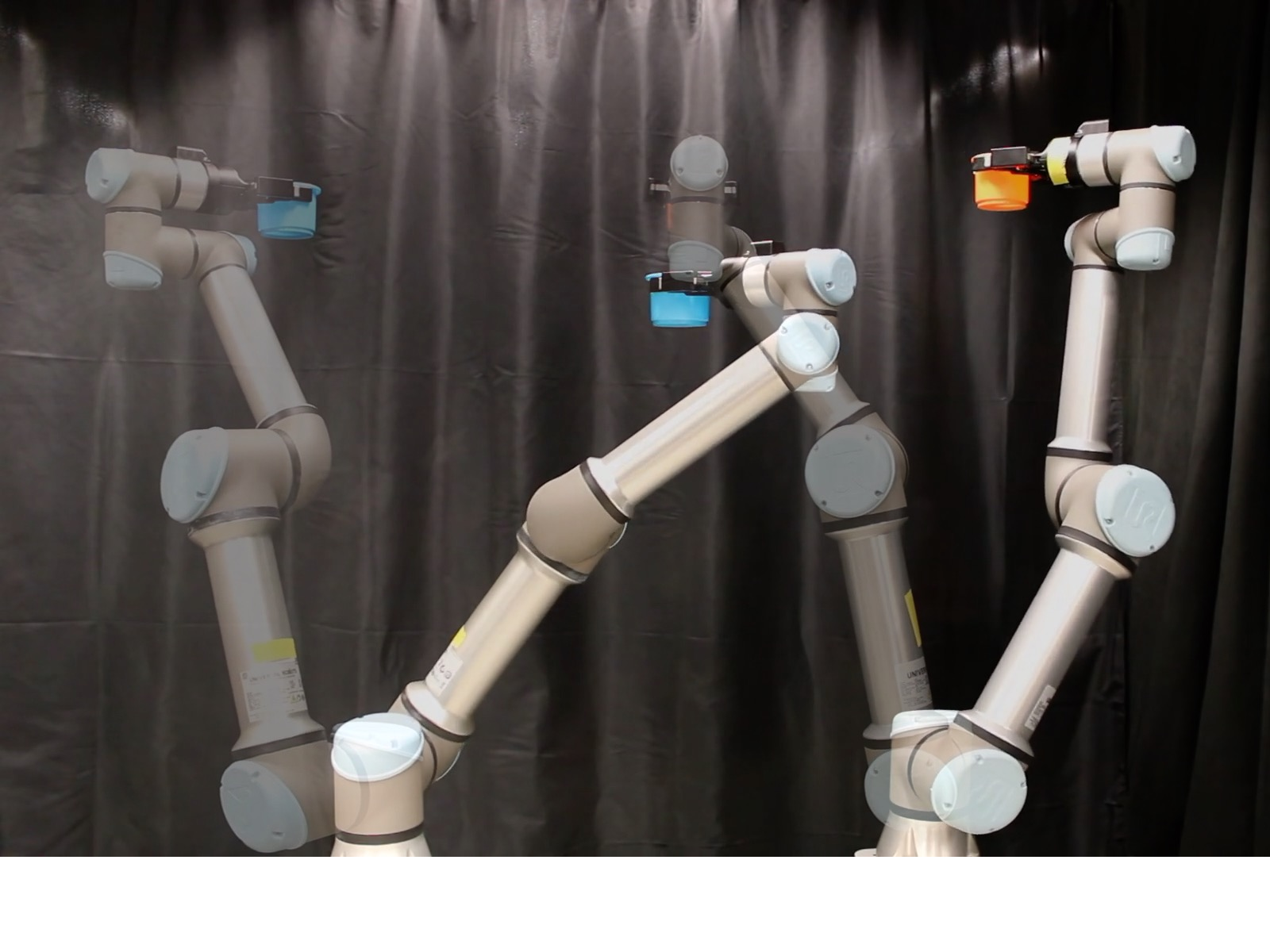}
\put(2,10){\color{white}\bfseries\small (4)}
\end{overpic}
\end{subfigure}\begin{subfigure}[t]{0.16\textwidth}
\begin{overpic}[width=\textwidth]{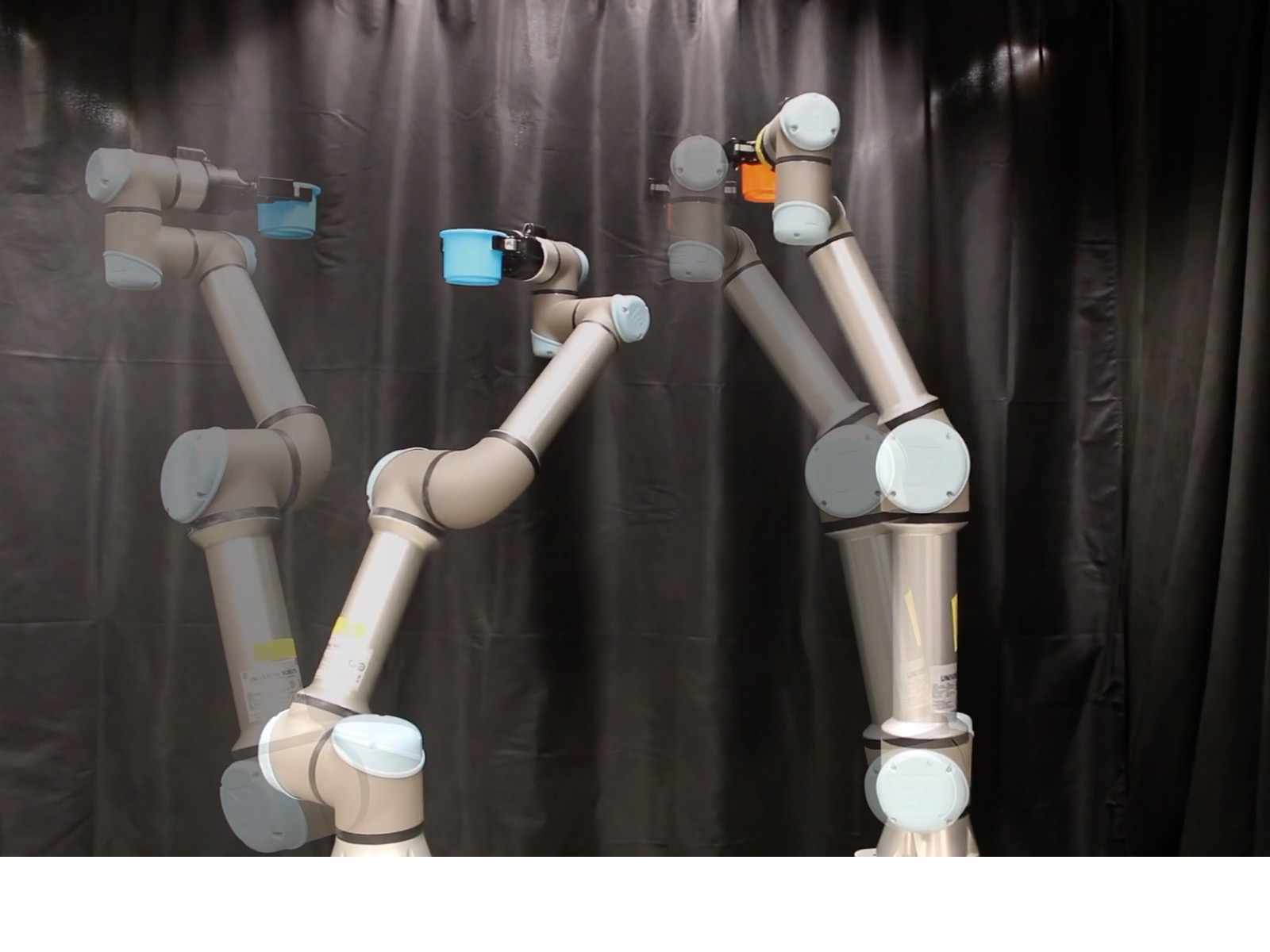}
\put(2,10){\color{white}\bfseries\small (5)}
\end{overpic}
\end{subfigure}\begin{subfigure}[t]{0.16\textwidth}
\begin{overpic}[width=\textwidth]{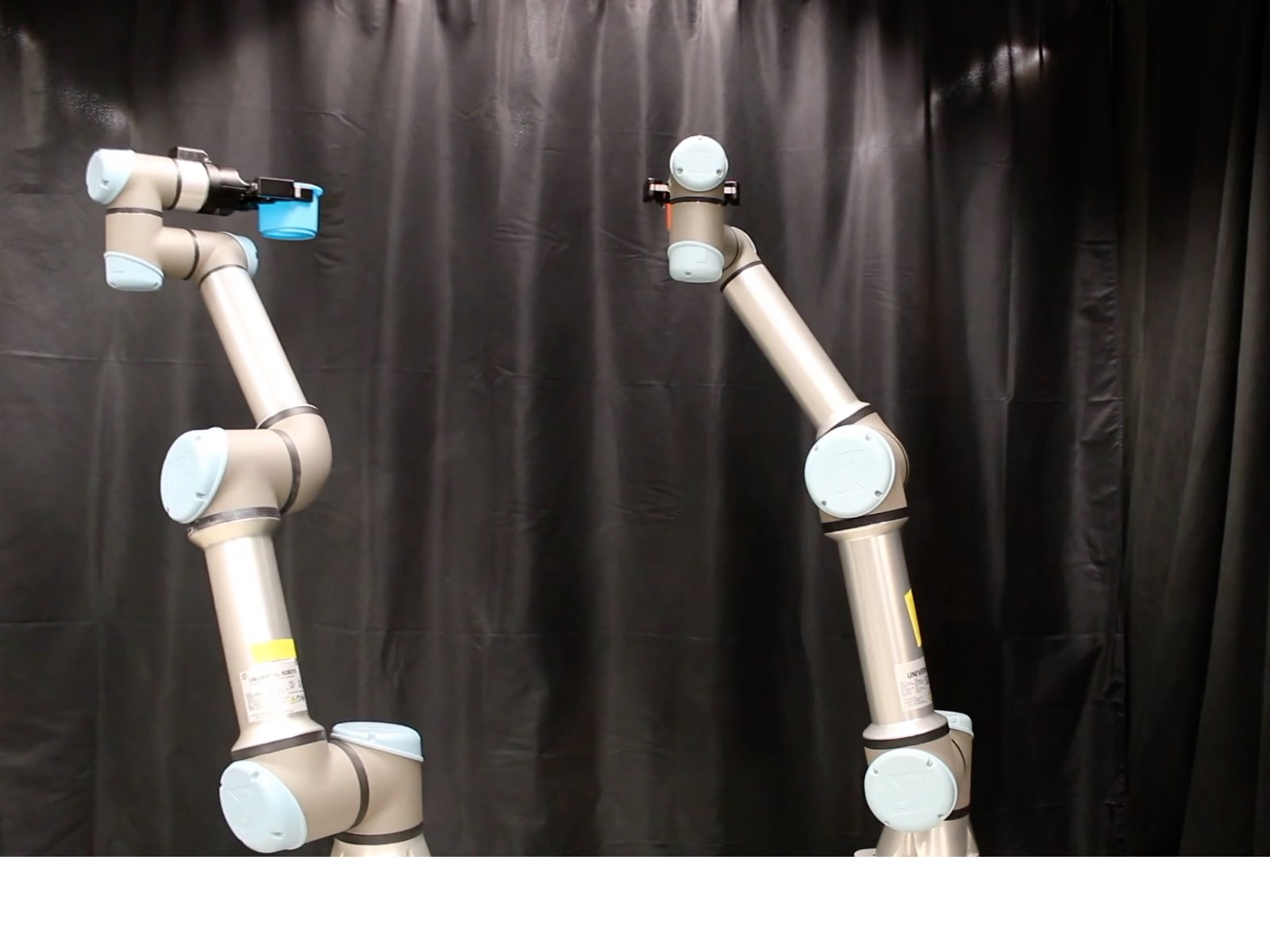}
\put(2,10){\color{white}\bfseries\small (6)}
\end{overpic}
\end{subfigure}
\caption{Our method, \methodname{}, successfully solving a real-world cup-holding task with two UR5 manipulators. Transparent models indicate the goal configurations. The sequence illustrates how the manipulators move from their initial states and avoid potential collisions to reach their goals.}
\label{fig:hardware}
\end{figure}

% \begin{figure}[t!]
% \centering
% \begin{subfigure}[t]{0.24\textwidth}
% \begin{overpic}[width=\textwidth]{figures/hardware/cropped-2.png}
% \put(2,10){\color{white}\bfseries\small (1)}
% \end{overpic}
% \end{subfigure}\begin{subfigure}[t]{0.24\textwidth}
% \begin{overpic}[width=\textwidth]{figures/hardware/cropped-3.png}
% \put(2,10){\color{white}\bfseries\small (2)}
% \end{overpic}
% \end{subfigure}
% \vskip -1.8ex
% \begin{subfigure}[t]{0.24\textwidth}
% \begin{overpic}[width=\textwidth]{figures/hardware/cropped-4.png}
% \put(2,10){\color{white}\bfseries\small (3)}
% \end{overpic}
% \end{subfigure}\begin{subfigure}[t]{0.24\textwidth}
% \begin{overpic}[width=\textwidth]{figures/hardware/cropped-5.png}
% \put(2,10){\color{white}\bfseries\small (4)}
% \end{overpic}
% \end{subfigure}
% \caption{An example of \methodname{} successfully solving a \textbf{cup-holding} task. The robots’ current configurations are shown in solid gray with their goals shown in transparent green. The sub-figures illustrate a single trial at different time steps, where the manipulators start from their initial configurations and get close to each other (a-b), adjust their configurations to avoid a potential collision (c-d), and then safely move towards their goals (e). The corresponding video demonstration is available in the supplementary materials.}
% \label{fig:cup}
% \end{figure}

In contrast to the above approaches, Hamilton-Jacobi reachability (HJR) offers a principled way to analyze safety and achieve dynamic optimal control of multi-agent systems. Recent advances have leveraged the strong approximation ability of neural networks to overcome the exponential scaling limitation of traditional HJR solvers and paved the way for various applications in safe MAMP \cite{bansal2020deepreach, fisac-rl2019, hsu2024isaacsiterativesoftadversarial, Hsu__2021, pmlr-v242-jeong24a, recentadvances, chen2025nehmoneuralhamiltonjacobireachability}. Despite these advances, existing approaches do not often consider task-specific manifold constraints, thus motivating our integration of constrained motion planning with HJR.

This paper proposes \methodname{}, \textbf{Ha}milton–Jacobi with \textbf{M}anifold constraints for \textbf{M}ulti-\textbf{A}gent \textbf{R}eachability, a framework that learns manifold-constrained HJR for decentralized multi-agent motion planning. Our approach extends learning-based HJR solvers (i.e., DeepReach \cite{bansal2020deepreach}) to learn reachability value functions on constraint manifolds, ensuring safety analysis under task-specific requirements. The value function is then incorporated into a receding-horizon trajectory planning framework to generate motion plans that both account for collision avoidance and respect task constraints. \methodname{} works as a multi-agent motion planner in a decentralized manner without assuming the control policies of other agents. Our contributions are summarized as follows:
\begin{enumerate}
    \item A theoretical extension to DeepReach \cite{bansal2020deepreach} that solves HJR under manifold constraints.
    %\qc{the derivation?}
    \item A decentralized multi-agent trajectory planner that generate motion plans that avoid collisions while respecting task-induced manifold constraints.
    \item A demonstration of our method on various constrained MAMP tasks, such as object-carrying, cup-holding, and doorway-crossing, to illustrate the applicability of our approach to realistic daily-life scenarios.
\end{enumerate}

Our results demonstrate that \methodname{} generalizes across a range of constrained MAMP tasks, generating decentralized motion plans that are both safety-aware and task-compliant. Figure \ref{fig:hardware} shows snippets of HAMMAR performing a real-world MAMP task of cup-holding under orientation constraints. Corresponding demonstration videos are provided in the supplementary materials.

\section{Related Work}

Decentralized multi-agent motion planning studies the problem of enabling multiple agents to independently generate motion plans that guide them from start to goal configurations without colliding with each other. Recent advances have seen promising results in scaling decentralized methods to high-dimensional systems such as robot manipulators. Model Predictive Control (MPC) and optimization-based methods \cite{Hartmann_2023, cooperative2020}  achieve decentralized collision avoidance by embedding safety constraints into the optimization, while reinforcement learning-based approaches \cite{ha2020learningdecentralizedmultiarmmotion, Lai_2025} extract coordinated motion policies from experience. However, these methods typically assume robots operate in the full configuration space and do not explicitly account for task-induced manifold constraints. Reactive methods based on potential functions \cite{rodriguez2024guaranteed, bakker2023multirobotlocalmotionplanning} enable local collision avoidance and can in principle be adapted to constrained settings, but their decisions are often myopic. In general, ensuring safety in dynamic environments requires systematically reasoning about the influence of manifold constraints on the robot's ability to avoid collisions and carefully generating corresponding motion plans.

Another line of research addresses the problem of constrained motion planning, which aims at finding collision-free paths under kinematic constraints. A representative example is AtlasRRT \cite{jaillet2012asymptotically}, which incrementally explores the constraint manifold using local charts. Extensions of sampling-based methods \cite{sampling-dual-arm, Kaczmarz, guo2024efficientmultirobotmotionplanning} have also demonstrated promising results in multi-agent cooperative tasks. However, these approaches generally emphasize offline planning and lack the ability to adapt online in dynamic or uncertain environments. Moreover, task constraints and inter-agent safety are often handled in centralized frameworks, limiting their applicability in decentralized  scenarios.

Hamilton-Jacobi reachability provides a complementary perspective to solve MAMP by characterizing safe sets and optimal control policies in multi-agent scenarios \cite{leung2020infusingreachabilitybasedsafetyassurance, chen2016multi, bansal2021provablesafemultivehicle}. Recently, neural HJR solvers have demonstrated promising results in scaling HJR analysis to higher-dimensional systems \cite{bansal2020deepreach, fisac-rl2019, feng2025bridging} and have enabled applications in safety-critical planning, including decentralized settings \cite{chen2025nehmoneuralhamiltonjacobireachability, pmlr-v242-jeong24a}. Inspired by this line of research, our method extends HJR to explicitly incorporate manifold constraints, enabling decentralized planning that is both safety-aware and task-feasible.
\section{Preliminaries and Background}

\subsection{Problem Setup}
% problem setup and assumptions
We investigate the problem of safe decentralized MAMP under manifold constraints, where agents are tasked to independently navigate toward their goals while avoiding collisions with other agents and adhering to their manifold task constraints. We assume that each agent can perfectly perceive the states of other agents and know their dynamics. However, we assume no knowledge of the control policies of other agents.

\subsection{Notation}
We denote the state (configuration) of a robot by $x \in \mathcal{X}$ where $\mathcal{X} \subseteq \mathbb{R}^{n_d}$ is its state (configuration) space with dimensionality $n_d$. We use $\mathcal{M} \subset \mathcal{X}$ to denote a manifold embedded in the ambient space $\mathcal{X}$ and use $T_x\mathcal{M}$ to denote the tangent space at $x\in M$,

When multiple agents are involved, we use superscripts to distinguish them by writing $x^i \in \mathcal{X}$ as the state of the $i$-th agent. We use $\langle\cdot, \cdot\rangle$ to denote a vector inner product. Other notations will be introduced and explained within the context of the discussion.

\subsection{Hamilton-Jacobi Reachability} \label{hjr}

Hamilton-Jacobi Reachability is a foundational method in optimal control theory, offering mathematical guarantees in computing the Backward Reachable Set (BRS), the set of initial states that can reach a target set given a specified time instance and system dynamics. 

Consider the optimal control problem of a system with dynamics $\Dot{x} = f(x, u)$ and a cost function $\texttt{cost}(x(t), u(t))$, where $u\in \mathcal{U}$ is the control input with  $\mathcal{U}$ being compact. The corresponding cost of a trajectory is then defined by
\begin{equation}
   J(t, x(t), u(t)) = l(x(T)) + \int_{t_0}^T \texttt{cost}(x(t), u(t))dt
\end{equation}
where $T$ is the terminal time and $l(x(T))$ is a bounded and Lipschitz-continuous function that computes the terminal cost. The value function $V: [0,T]\times \mathbb{R}^n \rightarrow \mathbb{R}$ 
\begin{equation}
    V(t,x(t)) = \min_{u(\cdot)}J(t, x(t), u(t)).
\end{equation}
then defines the optimal cost that can be achieved from state $x$ at time $t$. Applying the principle of dynamic programming yields the Hamilton-Jacobi Bellman (HJB) partial differential equation (PDE) to solve for $V(t,x)$
\begin{align}
   & \frac{\partial V}{\partial t} + \min_u\{\langle\nabla_x V(t, x, f(x,u))\rangle + \texttt{cost}(x, u)\} = 0 \\
   & \phantom{-------} V(T, x(T)) = l(x(T)).
\end{align}

In motion planning, the target set can be described by a level set $\mathcal{L} = \{x \mid l(x) \le 0\}$. The solution $V(t,x) \le 0$ if and only if the system can reach $\mathcal{L}$ by time $T$. The set $\mathcal{V}(t) = \{x \mid V(t,x)\le 0\}$ thus defines the backward reachable set. The same principle extends to target-avoiding or reach-avoid tasks, as well as multi-agent settings. We refer interested readers to \cite{mitchell2002application, fisac2014reachavoidproblemstimevaryingdynamics, bansal2017hamiltonjacobireachabilitybriefoverview} for further background on HJR.
\section{Proposed Method}

In this section, we introduce \methodname{}, which formulates a constrained HJR problem, solves it in a data-driven framework, and constructs a trajectory optimization planner that generates safe, task-feasible motions.

\subsection{Manifold-Constrained Hamilton-Jacobi Reachability}
Consider a dynamical system
$
    \Dot{x} = f(x,u), x\in \mathcal{X}\subseteq \mathbb{R}^{n_d}, u \in \mathcal{U}
$
 constrained by $n_c$ smooth equality constraints $C(x) = \begin{bmatrix}c_1(x) & c_2(x) & \cdots & c_{n_c}(x) \end{bmatrix}^T = 0$, with $0 < n_c < n_d$. These constraints $C:\mathcal{X}\rightarrow \mathbb{R}^{n_c}$ implicitly define a constrained configuration space (manifold) $\mathcal{M} = \{x \in \mathcal{X}\mid C(x)=0\}$ embedded in the ambient configuration space $\mathcal{X}$  (e.g., a vehicle constrained to a road curve, or a manipulator constrained to maintain an orientation). 

Since the trajectory of the system must remain on $\mathcal{M}$ rather than the full configuration space $\mathcal{X}$, reachability must be analyzed intrinsically on this manifold. In particular, the principle of dynamic programming for the value function over a short trajectory $V:[0,T]\times \mathcal{M}\to\mathbb{R}$ satisfies
\begin{align}
    & \phantom{==} V(t, x(t)) \nonumber\\
    &= \min_{u}\{V(t+\delta, x(t+\delta)) + \int_t^{t+\delta}\texttt{cost(}x(\tau),u(\tau)\texttt{)}d\tau\} \nonumber \\
    & \phantom{=========}\textbf{s.t. } x(\tau) \in \mathcal{M}\ \forall \tau \in [t, t+\delta]  \\
    & = \min_{u}\{V(t, x(t)) + \frac{\partial V}{\partial t}(t,x(t)) \cdot \delta + \langle \nabla V(t, x(t)), \frac{dx}{dt}\rangle \cdot \delta \nonumber \\ 
    & \phantom{=====} + \texttt{cost(}x(t),u(t)\texttt{)}\delta\}\nonumber\\
    & \phantom{=========}\textbf{s.t. } x(\tau) \in \mathcal{M}\ \forall \tau \in [t, t+\delta]  
\end{align}
with a first-order Taylor Expansion. Taking the limit of $\delta$ to zero yields a constrained form of the HJB PDE
\begin{align} \label{eqn:constrained-hjb}
    & \frac{\partial V (t,x)}{\partial t} + \min_u\{\langle \nabla V(t, x), f(x,u)\rangle + \texttt{cost(}x,u\texttt{)}\} = 0 \nonumber\\
    &\phantom{========} \textbf{s.t. }\Dot{C}(x) =  \begin{bmatrix}
         \langle \nabla c_1(x) , f(x,u)\rangle \\ 
         \langle \nabla c_2(x) , f(x,u)\rangle  \\
         \cdots \\
         \langle \nabla c_{n_c}(x) , f(x,u)\rangle 
    \end{bmatrix} = 0
\end{align}  
with the terminal condition $V(T,x) = l(x), x\in \mathcal{M}$. The corresponding Hamiltonian is then
\begin{align} \label{eqn:constrained-ham}
    & H_{\mathcal{M}}(t,x,p) = \min_u\{ \langle p, f(x,u)\rangle + \texttt{cost(}x,u\texttt{)}\} \nonumber\\
    &\phantom{==============} \textbf{s.t. }\Dot{C}(x) = 0
\end{align}
where the costate $p = \nabla V(t, x)$. 
Importantly, since $V$ is only defined on the manifold $\mathcal{M}$, the Euclidean gradient $\nabla V(t, x)$ is not globally meaningful. Therefore, when we compute the Hamiltonian and write $\nabla V$, we are referring to a directional derivative of the form $DV(t,x)[v] = \langle \nabla_\mathcal{M} V, v\rangle, \forall v\in T_x\mathcal{M}$ where $\nabla_\mathcal{M}V$ denotes a manifold gradient. We also assume satisfactions of the standard conditions for the HJR framework \cite{mitchell2002application} and that $\nabla c_i(x)$ is bounded $\forall i=1,\cdots, c_n\ \forall x\in \mathcal{M}$. Essentially, the equality constraints restrict admissible dynamics to the tangent bundle 
\begin{align}
    T\mathcal{M} = \{(x, v) \in (\mathcal{M}, \mathbb{R}^{n_d})\mid J_C(x) v = 0\}
\end{align}
where $J_C(x)$ is the Jacobian of the constraints, so that feasible trajectories evolve intrinsically on $\mathcal{M}$. As in the unconstrained setting, the resulting PDE must be understood in the viscosity sense. We note that a majority of the derivation follows the one in the unconstrained case and refer interested readers to \cite{lavalle2006planning, mitchell2002application} for more details.

\subsection{Solving Manifold-Constrained HJR with DeepReach}

This section discusses how we solve manifold-constrained HJR problems (i.e., Equation \ref{eqn:constrained-hjb}) as a safety constraint for multi-agent motion planning. In safe robot motion planning, HJR is often concerned with the Backward Reachable Tube (BRT) problem formulated by
\begin{align}\label{hjipde}
    \frac{\partial V(t, x)}{\partial t} &+ \min\{H(t, x, \nabla V(t,x)), 0\} = 0 \nonumber\\
    &V(T, x) = l(x).
\end{align}
where $H$ is the Hamiltonian of the system. For systems with disturbances, 
\begin{equation} \label{eqn:ham}
     H(t, x, \nabla V(t,x)) = \max_{u}\min_{d}\langle \nabla V(t,x), f(x,u,d)\rangle
\end{equation}
with $u\in \mathcal{U}$ being the control to the system and $d \in \mathcal{D}$ being the disturbance to the system. The inclusion of a non-anticipative disturbance component reflects a zero-sum differential game \cite{fisac2014reachavoidproblemstimevaryingdynamics} setting that well models the interaction between an agent and the other adversary agent(s), thus producing a safety value function for multi-agent motion planning. The solution ensures that by acting optimally, the agent can remain safe for the horizon $[t, T]$ given that $V(t,x) > 0$. In these cases, the boundary condition $l(x)$ is often chosen as a signed distance function that distinguish safe and unsafe states by its sign. 

DeepReach \cite{bansal2020deepreach} is a leading approach that solves Equation \ref{hjipde} using neural networks, overcoming the exponential scaling limitation of traditional methods. For a given training point $(t, x)$, it defines the loss function $L(t, x; \theta)$ as
\begin{equation}
    L(t, x; \theta) = L_1(t,x;\theta) + \lambda L_2(t, x;\theta)
\end{equation}
where
\begin{align}
    L_1(t, x; \theta) &= |V_\theta(t, x) - l(x)|\mathds{1}(t=T) \\
     L_2(t, x; \theta) &= |\frac{\partial V_\theta(t, x)}{\partial t} + \min\{0, H(t, x, \nabla V(t, x))\}|. \label{eqn:loss2}
\end{align}
The network weights $\theta$ are optimized to minimize the loss function. For further details on DeepReach, we refer interested readers to \cite{bansal2020deepreach}.

We now discuss how to extend this framework to manifold-constrained settings. Recall from Equations \ref{eqn:constrained-hjb} and \ref{eqn:constrained-ham} that the key difference is the restriction of system dynamics to the tangent bundle of the constraint manifold. This therefore requires replacing the unconstrained Hamiltonian in Equation \ref{eqn:loss2} with its constrained counterpart. While this generally entails solving a constrained optimization problem at each step and may lead to inefficiency in a training, there are important cases where a closed-form constrained Hamiltonian can be derived.

Consider a velocity-controlled system $\Dot{x} = f(x,u) = u$, $||u||_2 \le \Bar{u}$. This for example corresponds to a particle system controlled by a velocity input, up to a maximum speed $\Bar{u}$. The manifold-constrained Hamiltonian of this system in a HJB formulation is then
\begin{align}
    &\phantom{==} H_{\mathcal{M}}(t, x, \nabla V(t,x)) \\
    &= \min_{u}\langle \nabla V(t,x), u\rangle \phantom{=} \textbf{s.t. }\Dot{C}(x) = 0, ||u||_2 \le \Bar{u} \label{plug-in}\\
    &= \min_{u}\langle \nabla V(t,x), u\rangle \phantom{=} \textbf{s.t. } J_C(x)u = 0, ||u||_2 \le \Bar{u}
\end{align}
where $J_C(x) \in \mathbb{R}^{n_c\times n_d}$ is the Jacobian of the constraint function $C(x)$. This can be simplied as 
\begin{align}
    &\phantom{==} H_{\mathcal{M}}(t, x, \nabla V(t,x)) \\ &= \min_{u} \langle \nabla V(t,x), P(x) u\rangle \phantom{=} \textbf{s.t. } ||u||_2 \le \Bar{u} \\
    &=  -\Bar{u}\ \|P(x) \nabla V(t,x) \|_2
\end{align}
with the minimizer
\begin{equation} \label{eqn:u-min}
    u^* = -\Bar{u}\frac{P(x) \nabla V(t,x)}{\|P(x) \nabla V(t,x) \|_2}
\end{equation}
and the projection matrix 
\begin{equation}
    P(x) = \mathbb{I} - J_C(x)^T(J_C(x)J_C(x)^T)^{-1}J_C(x).
\end{equation}
 Here $P(x)$ projects vectors onto the nullspace of $J_C(x)$, ensuring that control inputs lie in the tangent space of the manifold. We notes that $J_C(x)$ is required to be full row rank so that $J_CJ_C^T$ is invertible. Although derived for a velocity-controlled system, such closed-form solution to constrained Hamiltonian can be similarly obtained for linear control-affine systems. % We refer interested readers to Appendix \ref{} for more details.

\subsection{Trajectory Optimization with Neural HJR }

In this section, we extend NeHMO \cite{chen2025nehmoneuralhamiltonjacobireachability} to manifold-constrained scenarios to generate motion plans that guide an agent toward its goal, satisfy the specified manifold constraints, and avoid collisions in the presence of other agents. At each planning step, the computed trajectory must remain on the manifold $\mathcal{M}$, making sure the value function $V_\theta$ is well-defined. Moreover, the  value function $V_\theta(t, x)$ is required to stay above a specified safety threshold $\epsilon$ to achieve collision avoidance. These requirements lead to a receding-horizon trajectory optimization problem \Opt formulated as follows:
\begin{align} \label{opt}
    & \min_{u\in \mathcal{U}} \int_{0}^{t_{\text{plan}}} \texttt{cost(}(x^1(t), u, x^1_{\text{goal}}\texttt{)}\ dt \\
    \textbf{s.t.} & \hspace{0.15cm} x^{1}(t+\Delta t) = f_d(x^1(t), u)\nonumber\\
    & \hspace{0.15cm} x^{i}(t+\Delta t) = f_d(x^i(t), d^{i})\nonumber\\
    & \hspace{0.15cm} C(x(t_{\text{plan}})) = 0 \nonumber \\
    & \hspace{0.15cm} \tau = T - t_{\text{plan}} \nonumber\\
    & \hspace{0.15cm} d^{i} = \arg\min_{d\in \mathcal{D}}\max_{u\in \mathcal{U}}\langle \nabla V_\theta(\tau,x^1(t), x^i(t)), f(x(t),u,d) \rangle \nonumber\\
    & \hspace{0.15cm} V_\theta(T - t_{\text{safe}}, x^{1}(t_{\text{plan}}), x^{i}(t_{\text{plan}})) > \epsilon \nonumber \\
    & \phantom{====================} \text{ for } i=2,\cdots,m \nonumber
\end{align}
Here, $m$ is the number of agents in the environment, $\epsilon$ is a safety buffer, and $f_d(x^i(t), u)$ is the discrete-time model propagation function that computes the next state of an agent given its current state and a control input. The manifold constraints ensure that the agent’s next state lies on $\mathcal{M}$, while the HJR value function is computed pair-wise with every other agent to enforce the planned trajectory to be collision-free under worst-case actions from other agents.

The solution of the optimization provides the control input for the next planning horizon. To ensure safety, we adopt a fail-safe strategy: if the optimization fails to solve, the agent falls back to a conservative control given by the most collision-evading input computed similar to Equation \ref{eqn:u-min}. During execution, the agent follows its current plan while simultaneously computing the next one, enabling real-time adaptation to environmental changes and the states of other agents.

%\subsection{Decentralized Multi-Agent Motion Planner}
%\qc{do we still need this subsection?}
%Finally, Algorithm \ref{alg:online_planning} summarizes the online planning procedure. At each planning step, the agent acquires the current configurations of all agents in the environment and formulates the optimization problem \optref. The computed solution provides the control input for the next planning horizon. To ensure safety, we incorporate a fail-safe strategy: if the optimization problem cannot be solved within the allocated time, the agent executes a conservative fallback control designed to avoid collisions while maintaining feasibility with respect to the manifold constraints.
%The agent executes its current motion plan continuously while simultaneously computing the next horizon plan. This overlapping planning-execution scheme enables real-time adaptation to changes in the environment and the states of other agents.

%TODO: FAILSAFE
%\input{algorithms/planning}
\section{Experiments and Validation} \label{sec:exp}

This section presents two sets of analyses to evaluate our proposed \methodname{} framework. First, we compare the HJR value functions computed with and without the manifold-constrained formulation on a 2-dimensional particle system. Second, we evaluate \methodname{} as a motion planner in three representative MAMP tasks that must operate under manifold constraints. In the object-carrying task (Fig. \ref{fig:object}), sub-teams of robots operate in a decentralized manner to transport a large object while respecting closed-loop manifold constraints.  In the cup-holding task (Fig. \ref{fig:cup}), multiple robots independently transport their cups while keeping it upright and avoiding collisions with one another. Finally, in the doorway-crossing task (Fig. \ref{fig:doorway}), we extend the classical doorway problem to high-DOF manipulators, requiring them to pass through a doorway while maintaining several end-effector constraints. All experiments are conducted using UR5 robots. For each problem setup, we generate 100 test cases with distinct start and goal configurations for the multi-agent system.

We benchmark \textbf{\methodname{}} against the following baselines: (i) \textbf{\Opt w/o HJR}, an ablation wihout safety constraints, (ii) \textbf{AtlasRRT (decentralized)}, and (iii) \textbf{AtlasRRT (centralized)}. \textbf{AtlasRRT (decentralized)} runs AtlasRRT for each robot independently to generate manifold-constrained waypoints that connect them respectively from start to goal configurations. This baseline serves as a reference where planned paths are not coordinated by a trajectory planner, to highlight the necessity of a trajectory planner that reacts to the environment in real-time. \textbf{AtlasRRT (centralized)} plans the motion of all manipulators in a centralized manner. This baseline has the privilege to globally control the motion of all robots to avoid collisions, however at the cost of higher dimensionality. This baseline serves as a reference of the task difficulties. Evaluation of the MAMP experiments is based on the following metrics: 
\begin{itemize}
    \item Success Rate (\textbf{SR\%}): The ratio of  trials where all agents reach their goals without any collision.
    \item Collision Rate (\textbf{CR\%}): The ratio of trials where collisions happen at any time.
    \item Planning Time (\textbf{Time}): The mean time taken for each agent to generate a plan.
    \item Path Length (\textbf{PL}): The average path length across successful trials. 
\end{itemize}

All experiments are conducted on a desktop with Intel(R) Xeon(R) w5-2455X CPUs and an NVIDIA GeForce RTX 4070 GPU. The optimization problem \optref is solved with \texttt{IPOPT} \cite{wachter2006implementation}. The multi-arm environment is adapted from \cite{Michaux-SPARROWS-RSS-24} using \texttt{trimesh} \cite{trimesh} for collision detection. We assign a 5-second time limit to \textbf{AtlasRRT (decentralized)} and a 20-second time limit to \textbf{AtlasRRT (centralized)}.

\subsection{2D Circle-Constrained Particle} \label{sec:exp-2d}
This section analyzes the solution quality of the learned value function on a 2-dimensional circle-constrained particle system. As an ablation, we compare the BRS obtained from the value functions learned with and without the manifold-constrained formulation. 

We consider a 2D velocity-controlled particle system with state $x = (x_1, x_2) \in \mathbb{R}^2$ and  dynamics $\Dot{x} = f(x,u) = u \in \mathbb{R}^2$, $\|u\|_2 \le 1$. The  particle is constrained to a circle of radius 0.5 centered at the origin, thus defining the manifold constraint $c(x) = \|x\|_2 -0.5 = 0$. We now study a goal-reaching reachability problem with the goal located at $x_{\text{goal}} = (0.5, 0.0)$ and terminal condition $V(T, x) = l(x) = \|x - x_{\text{goal}}\|_2$. This corresponds to a geodesic distance problem and gives rise to the ground truth BRS $\mathcal{V}_{\text{gt}}(t) = \{(x_1, x_2) \in \mathcal{M} \mid \frac{1}{2}\arccos(2x_1) - (T - t) \le 0\}$. 

\begin{figure}[htbp]
    \centering
    \includegraphics[width=0.98\linewidth]{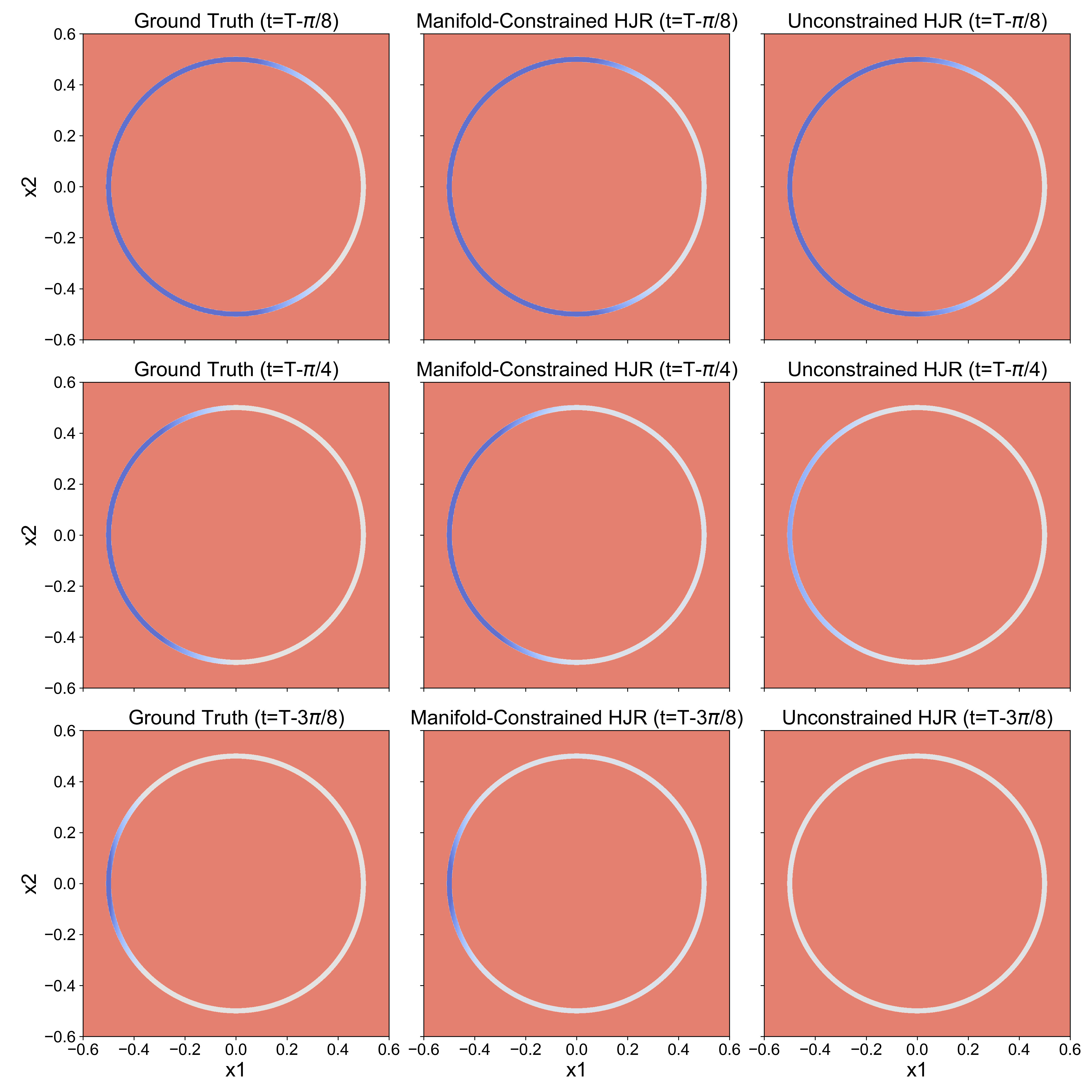}
    \caption{A comparison of the BRS obtained from the ground truth (left), the manifold-constrained HJR value function (middle), and the unconstrained HJR value function (right), each shown at three time slices. Backward reachable regions appear in white, backward unreachable regions in blue, and off-manifold regions in orange. The BRS obtained with the constrained formulation closely aligns the ground truth, while the unconstrained one over-approximates the BRS.}
    \label{fig:brs-comparison}
\end{figure}

Figure \ref{fig:brs-comparison} illustrates the BRS obtained from the value functions trained with and without the constrained formulation. As the time horizon increases from $\pi/8$ to $3\pi/8$, the BRS computed without the constrained formulation progressively over-approximates the BRS. For a quantitative analysis, we treat BRS prediction as a binary classification problem and report the two value functions' accuracy, recall, precision, and F-1 scores in Table \ref{tab:brs-comparison}. The unconstrained version over-approximates the BRS, achieving high recall by capturing all true positives. However, this over-approximation also introduces false positives, lowering its precision. In contrast, the constrained version closely matches the ground truth and achieves high performance across all metrics.

\begin{table}[htbp]
    \centering
    \begin{adjustbox}{width=0.9\columnwidth,center}
    \begin{tabular}{cccccc}
    \toprule
    Time & Manifold & Acc.\,$\uparrow$ & Rec.\,$\uparrow$ & Prec.\,$\uparrow$ & F-1\,$\uparrow$ \\\midrule 
    \multirow{2}{*}{$ T - \frac{\pi}{8}$} & w/ & \textbf{99.4} & 99.6 & \textbf{98.2} & \textbf{98.9} \\
    &  w/o & 97.4 & \textbf{100.0} & 90.7 & 95.1 \\\midrule
    \multirow{2}{*}{$ T - \frac{\pi}{4}$} & w/ & \textbf{99.5} & 99.3 & \textbf{99.7} & \textbf{99.5} \\
    & w/o & 89.7 & \textbf{100.0} & 83.0 & 90.7 \\\midrule
    \multirow{2}{*}{$ T - \frac{3\pi}{8}$} & w/ & \textbf{99.5} & 99.4 & \textbf{99.9} & \textbf{99.7} \\
    & w/o & 75.0 & \textbf{100.0} & 75.0 & 85.7 \\
    \bottomrule
    \end{tabular}
    \end{adjustbox}
    \caption{The accuray, recall, precision, and F-1 score of the BRS prediction obtained from the learned value functions with and without the manifold constraint.}
    \label{tab:brs-comparison}
\end{table}
\subsection{Object-Carrying UR5 Manipulators}
This section evaluates \methodname{} as a multi-agent motion planner in a factory-inspired dual-arm manipulation scenario. The setup consists of two UR5 leader manipulators placed face-to-face, each with an additional follower manipulator positioned at a distance. The followers %are sufficiently far from the leaders to safely ignore collisions, and they 
match the motion of their corresponding leaders to carry an object collaboratively under closed-loop manifold constraints.

\begin{figure*}[t!]
\centering
\begin{subfigure}[t]{0.195\textwidth}
\includegraphics[width=1\textwidth]{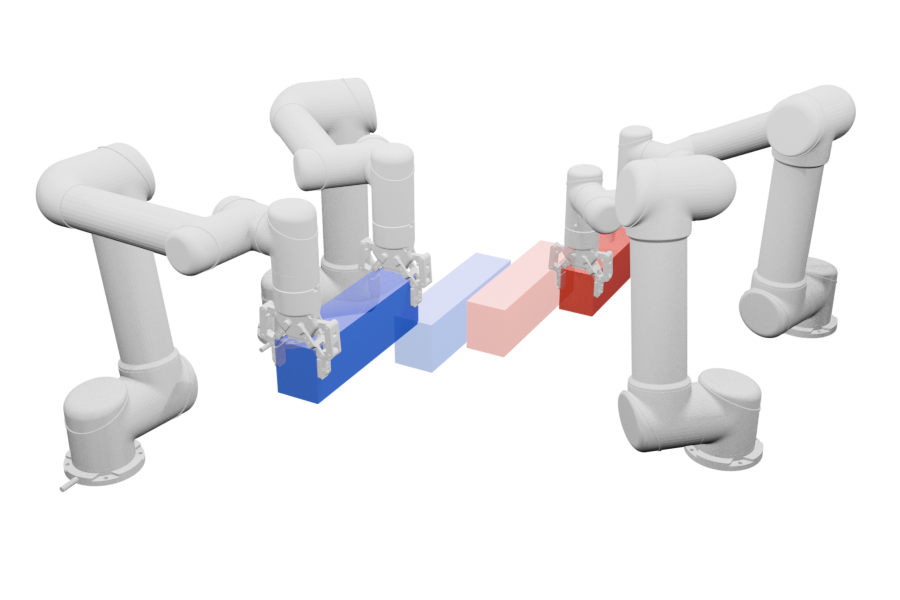}
\vskip -4ex
\caption{}
\end{subfigure}
\begin{subfigure}[t]{0.195\textwidth}
\includegraphics[width=1\textwidth]{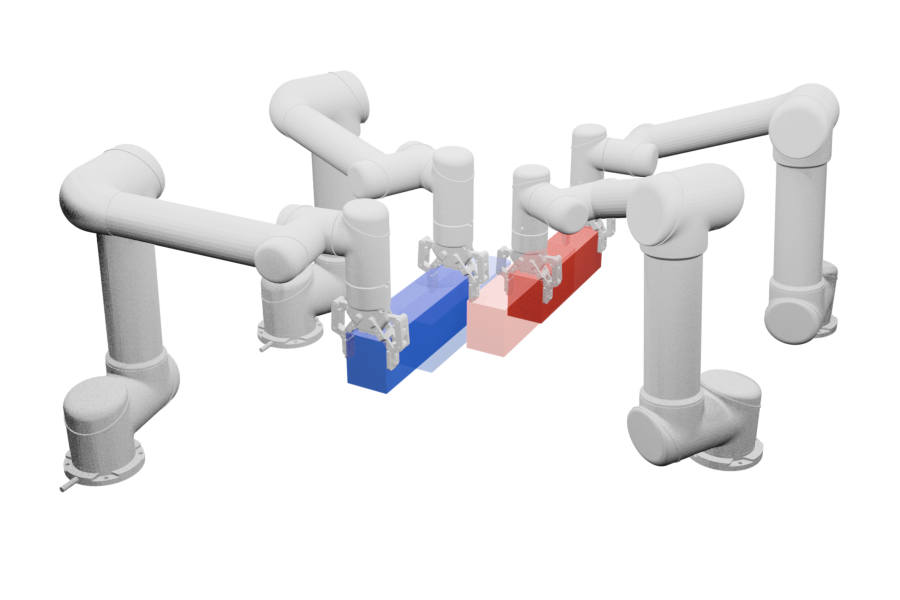}
\vskip -4ex
\caption{}
\end{subfigure}
\begin{subfigure}[t]{0.195\textwidth}
\includegraphics[width=1\textwidth]{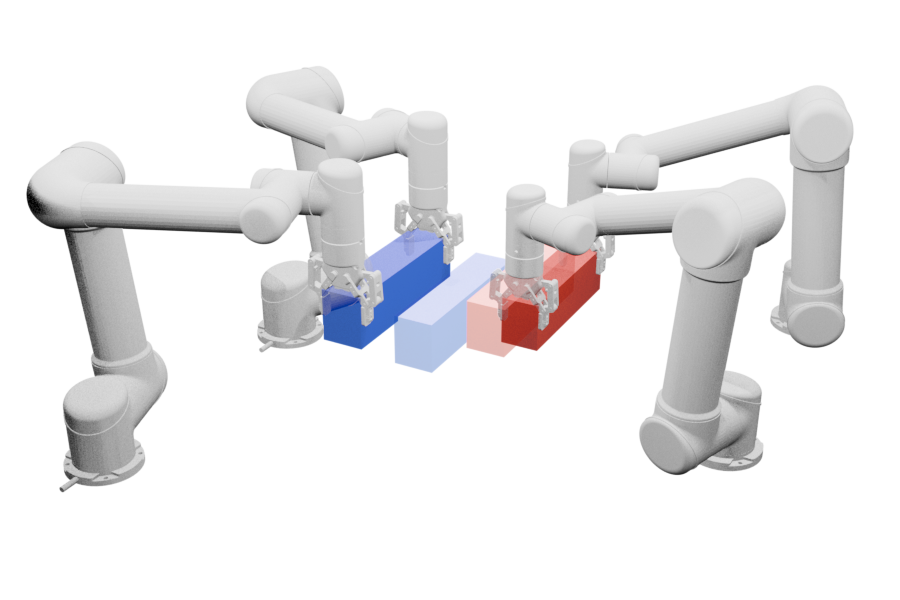}
\vskip -4ex
\caption{}
\end{subfigure}
\begin{subfigure}[t]{0.195\textwidth}
\includegraphics[width=1\textwidth]{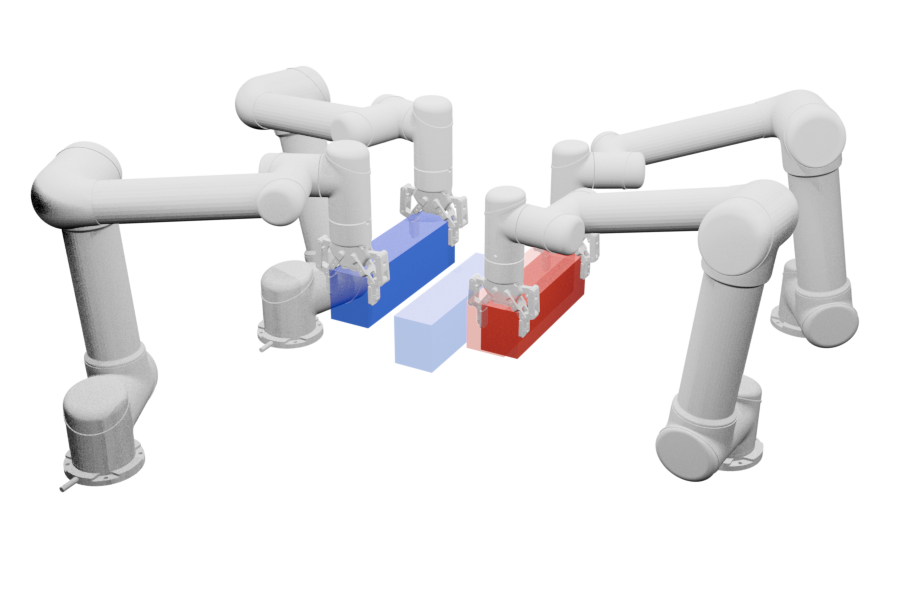}
\vskip -4ex
\caption{}
\end{subfigure}
\begin{subfigure}[t]{0.195\textwidth}
\includegraphics[width=1\textwidth]{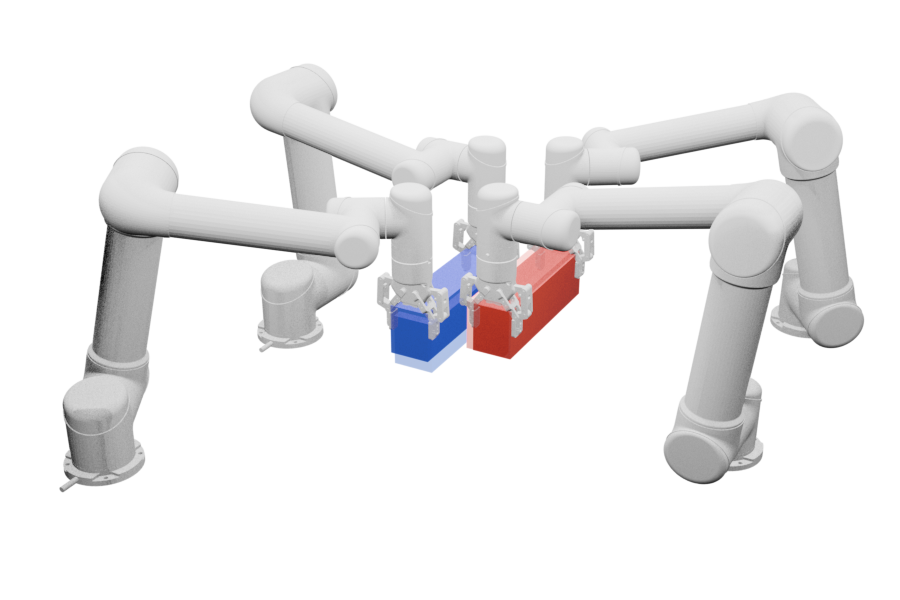}
\vskip -4ex
\caption{}
\end{subfigure}
\caption{An example of \methodname{} successfully solving an \textbf{object-carrying} task. The robots’ current configurations are shown in solid gray, while each object is shown in red or blue, with transparent counterparts indicating their respective goal positions. The sub-figures illustrate a single trial at different time steps, where the manipulators start from their initial configurations and get close to each other (a-b), adjust their configurations to avoid a potential collision (c-d), and then safely move towards their goals (e). The corresponding video demonstration is available in the supplementary materials.}
\label{fig:object}
\end{figure*}

We simulate a task where the leader robots aim to transport their respective objects to a shared workspace while avoiding collisions with each other. Both the initial positions of the objects and their goal positions in the shared workspace are randomized for each trial to vary task configurations. All manipulators are constrained to maintain a fixed end-effector orientation, introducing a manifold constraint on the system $C(x) = \texttt{orientation}(x) = R_{\text{desired}}$. This constraint simulates industrial requirements where the object must remain upright or aligned during handling. Figure \ref{fig:object} illustrates an example of the setup.

\begin{figure*}[t!]
\vskip -4ex
\centering
\begin{subfigure}[t]{0.195\textwidth}
\includegraphics[width=1\textwidth]{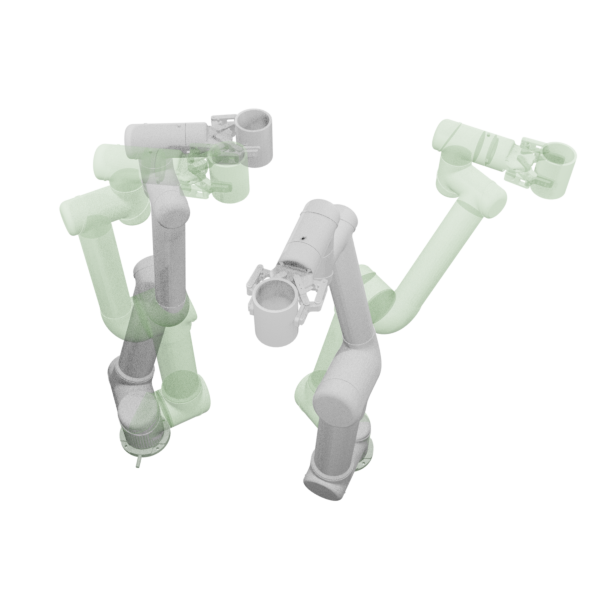}
\vskip -4ex
\caption{}
\end{subfigure}
\begin{subfigure}[t]{0.195\textwidth}
\includegraphics[width=1\textwidth]{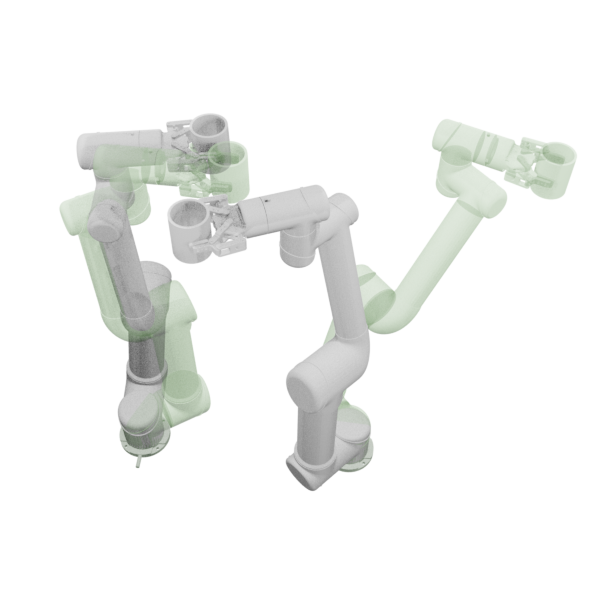}
\vskip -4ex
\caption{}
\end{subfigure}
\begin{subfigure}[t]{0.195\textwidth}
\includegraphics[width=1\textwidth]{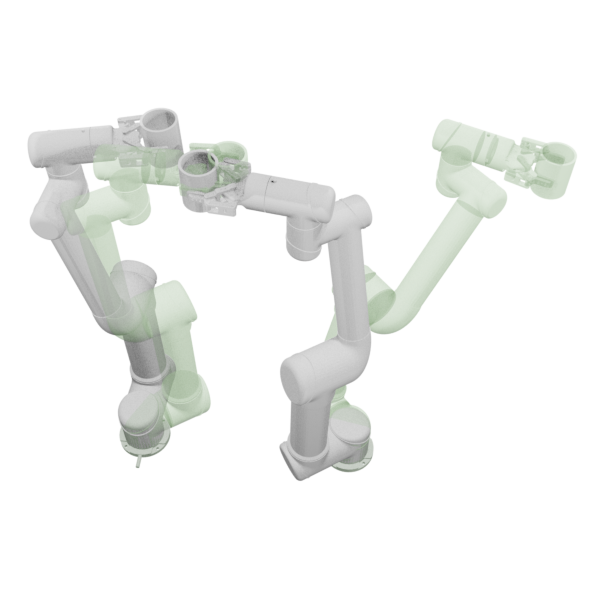}
\vskip -4ex
\caption{}
\end{subfigure}
\begin{subfigure}[t]{0.195\textwidth}
\includegraphics[width=1\textwidth]{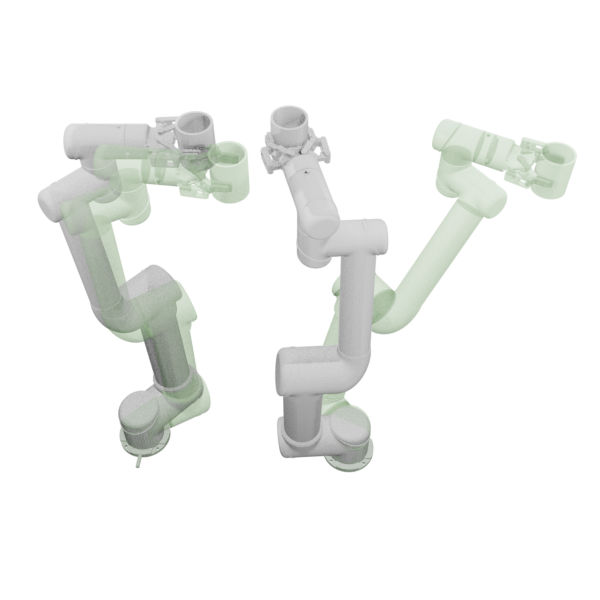}
\vskip -4ex
\caption{}
\end{subfigure}
\begin{subfigure}[t]{0.195\textwidth}
\includegraphics[width=1\textwidth]{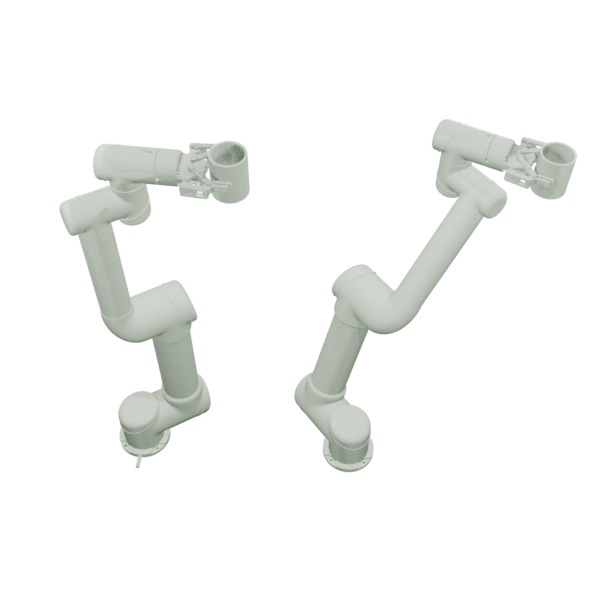}
\vskip -4ex
\caption{}
\end{subfigure}
\caption{An example of \methodname{} successfully solving a \textbf{cup-holding} task. The robots’ current configurations are shown in solid gray with their goals shown in transparent green. The sub-figures illustrate a single trial at different time steps, where the manipulators start from their initial configurations and get close to each other (a-b), adjust their configurations to avoid a potential collision (c-d), and then safely move towards their goals (e). The corresponding video demonstration is available in the supplementary materials.}
\label{fig:cup}
\end{figure*}

We solve for the leaders' motion to ensure collision avoidance and adherence to the manifold, while the followers match the leaders’ trajectories using inverse kinematics to maintain object stability. Table \ref{tab:object} summarizes the experiment result of the object-carrying task. \textbf{\methodname{}} outperforms the baselines both in achieving success and in avoiding collisions. \textbf{AtlasRRT (decentralized)} performs similarly with \textbf{\Opt w/o HJR}, highlighting the need of a trajectory planner in dynamic environments. When the two leader manipulators are jointly considered in a centralized manner, \textbf{AtlasRRT (centralized)} achieves more successes with zero collisions. However, the high-dimensionality of the problem prevents it from either frequently finding a valid solution or solving the problem in real-time.  

\begin{table}[htbp]
    \centering
\begin{adjustbox}{width=0.98\columnwidth,center}
    \begin{tabular}{ccccc}
    \toprule
    Methods & SR\% $\uparrow$ & CR\% $\downarrow$  & Time [s] $\downarrow$ & PL [rad] $\downarrow$ \\\midrule  
    \methodname{} (ours) &  \textbf{85} & \textbf{0} & 0.16 ± 0.15 & 2.7 ± 1.3 \\ 
    \Opt w/o HJR & 68 & 32 & \textbf{0.063 ± 0.023} & \textbf{2.1 ± 0.6}  \\ 
    AtlasRRT (decentralized) & 68 & 32 & 0.22 ± 0.14 & 2.4 ± 0.6 \\
    AtlasRRT (centralized) & 75 & \textbf{0} & 7.6 ± 8.6 & 2.5 ± 0.6 \\
    % AtlasRRT (joint, 100) & 72 & 27 & 16 ± 32 & 2.4 ± 0.7 \\
    \bottomrule
    \end{tabular}

\end{adjustbox}
    \caption{The success rate, collision rate, planning time, and mean path length of each method on the \textbf{object-carrying} task. Best results for each metric are shown in bold.}
    \label{tab:object}
\end{table}

\subsection{Cup-Holding UR5 Manipulators}
We further test our framework on a goal-reaching task in which multiple manipulators aim to reach their respective goal poses without colliding. Each manipulator holds a cup that must stay upright throughout the motion, introducing a manifold constraint on the end effector’s orientation in z-direction
$C(x) = \texttt{orientation}_{\texttt{z}}(x) = [0\ 0\ 1]^T$. Unlike the factory manipulation scenario, this task emphasizes pose-to-pose motion planning and involves longer-range motion, making collision-free path-finding more challenging. Figure \ref{fig:cup} illustrates an example of the setup.

\begin{wrapfigure}{R}{0.23\textwidth}
\centering
\vskip -3.5ex
\includegraphics[width=0.23\textwidth]{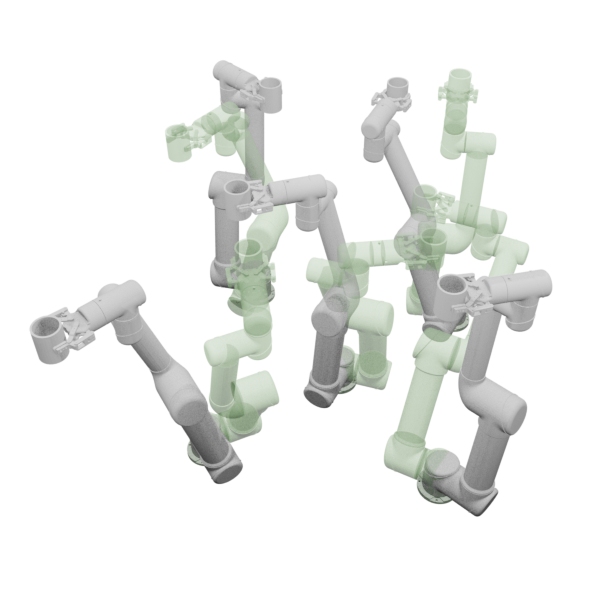}
\vskip -3ex
\caption{An example of the \textbf{five-UR5 cup-holding} task. Start configurations are shown in solid gray with goals shown in transparent green.}
\vskip -2ex
\label{fig:five}
\end{wrapfigure}

Table \ref{tab:cup} reports the results of
the cup-holding experiment with two manipulators. \textbf{\methodname{}} outperforms the decentralized baselines both in achieving success and in avoiding collisions. As the required range of motion increases compared to the object-carrying task, \textbf{AtlasRRT (centralized)} starts to struggle to find a path due to the necessity to adequately sample a high-dimensional manifold. For \textbf{real-world validation}, we evaluated our method on five challenging cup-carrying tasks with varying start and goal configurations. In all cases, \methodname{} executed successfully, achieving an average planning time of $0.10 \pm 0.051$ seconds per step and a mean path length of $11 \pm 3.3$. A visualization of the hardware experiment is shown in Figure \ref{fig:hardware}, with videos provided in the supplementary materials.

\begin{figure*}[ht]
\vskip -3ex
\centering
\begin{subfigure}[t]{0.245\textwidth}
\includegraphics[width=1\textwidth]{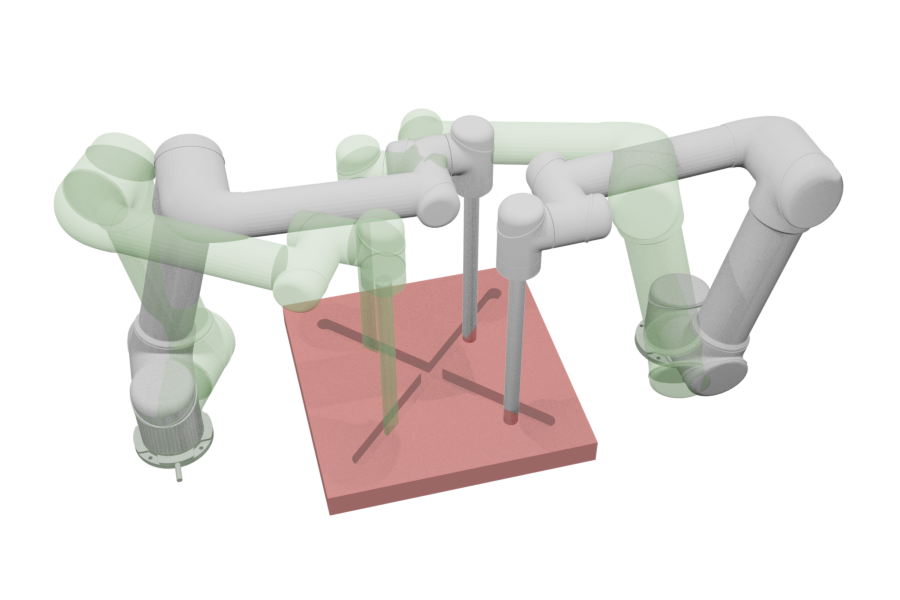}
\vskip -4ex
\caption{}
\end{subfigure}
\begin{subfigure}[t]{0.245\textwidth}
\includegraphics[width=1\textwidth]{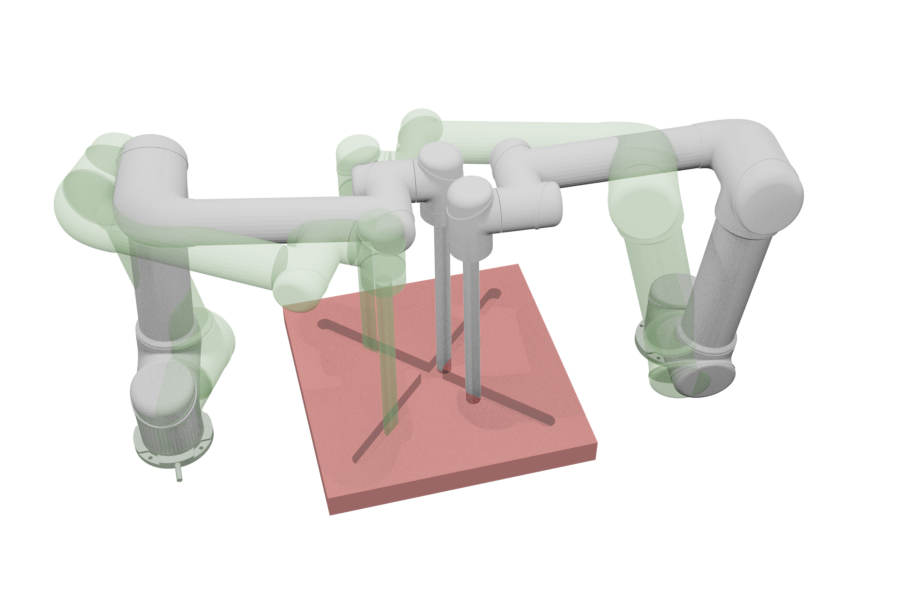}
\vskip -4ex
\caption{}
\end{subfigure}
\begin{subfigure}[t]{0.245\textwidth}
\includegraphics[width=1\textwidth]{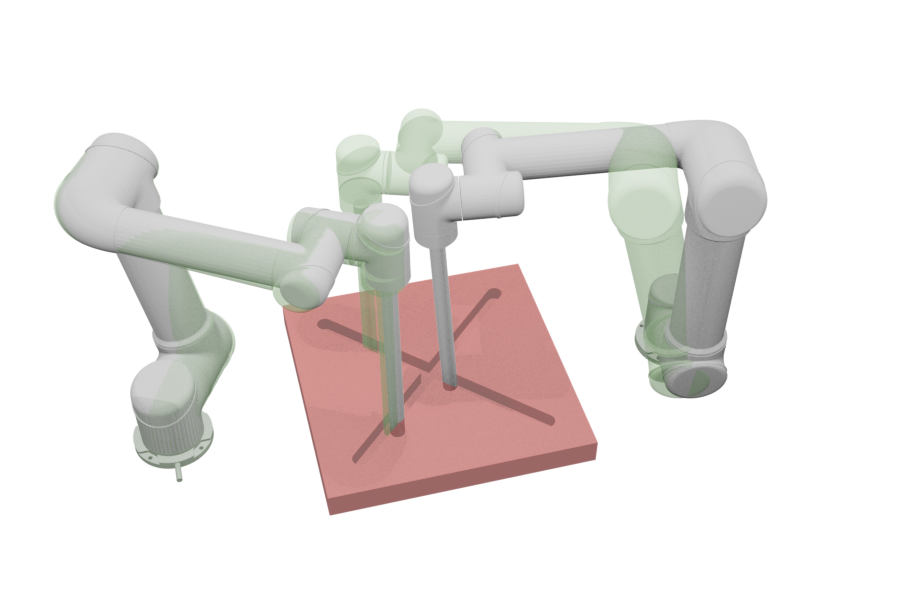}
\vskip -4ex
\caption{}
\end{subfigure}
\begin{subfigure}[t]{0.245\textwidth}
\includegraphics[width=1\textwidth]{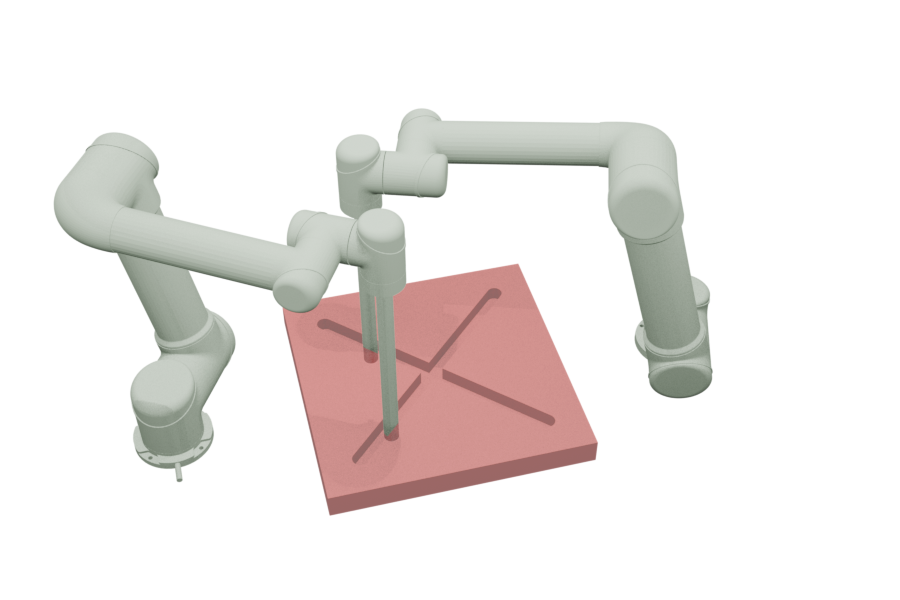}
\vskip -4ex
\caption{}
\end{subfigure}
\caption{An example of \methodname{} successfully solving a \textbf{doorway-crossing} task. The robots’ current configurations are shown in solid gray with their goals shown in transparent green. The sub-figures illustrate a single trial at different time steps, where the robots start from their initial configurations and get close to each other (a-b), wait until clearance (c), and then safely move towards their goals (d). The corresponding video demonstration is available in the supplementary materials.}
\label{fig:doorway}
\end{figure*}

To further assess the scalability of \textbf{\methodname{}}, we extend the cup-holding task to involve more agents. In these experiments, we surround a manipulator with two and four additional UR5 manipulators, thereby ensuring that each added agent meaningfully increases task difficulty. Planning time is therefore only measured on the surrounded manipulator, which faces the most challenging coordination requirement. An example configuration of the five-UR5 cup-holding is illusrated in Figure \ref{fig:five}. With three UR5 manipulators, \textbf{\methodname{}} achieves $78/100$ successful trials with 2 collisions, taking on average $0.12 \pm 0.13$ seconds per planning step. When scaled up to five manipulators, it records $57/100$ successes and 4 collisions while maintaining a reasonable planning time of $0.17 \pm 0.18$ seconds. In comparison, \textbf{AtlasRRT (centralized)} achieves fewer than 40 successes in the three-manipulator setting and fails to return any valid trajectory within the 20-second limit for five manipulators. The lower success rates in this setting reflect the higher risk of entanglement and tighter collision-avoidance constraints due to the presence of more agents. Still, \methodname{} is able to maintain safety to a large extent with efficiently scaled planning time. A video example of the five-manipulator setup is included in the supplementary materials.

\begin{table}[htbp]
    \centering
\begin{adjustbox}{width=0.98\columnwidth,center}
    \begin{tabular}{ccccc}
    \toprule
    Methods & SR\% $\uparrow$ & CR\% $\downarrow$  & Time [s] $\downarrow$ & PL [rad] $\downarrow$ \\\midrule  
    \methodname{} (ours) &  \textbf{82} & 2 & 0.10 ± 0.087 & 7.7 ± 3.1 \\ 
    \Opt w/o HJR & 67 & 33 & \textbf{0.056 ± 0.010} &  \textbf{6.8 ± 2.4} \\ 
    AtlasRRT (decentralized) & 67 & 33 & 1.9 ± 1.7 & 7.0 ± 2.4 \\
    AtlasRRT (centralized) & 60 & \textbf{1} & 15 ± 7.4 & \textbf{6.8 ± 2.2} \\
    % AtlasRRT (joint, 100) & 73 & 13 & 30 ± 39 & 7.1 ± 2.4 \\
    \bottomrule
    \end{tabular}

\end{adjustbox}
    \caption{The success rate, collision rate, planning time, and mean path length of each method on the \textbf{cup-holding} task. Best results for each metric are shown in bold. }
    \label{tab:cup}
\end{table}

\subsection{Doorway-Crossing for UR5 Manipulators}

Finally, we evaluate our method on a doorway problem designed to involve various manifold constraints. In this task, two UR5 manipulators are equipped with sticks and must coordinate to cross a doorway defined by grooves on a block. This geometry introduces narrow passages and tight clearances, making collision avoidance particularly challenging. 

The task enforces several manifold constraints on the system. First, the stick must be held upright such that  $c_1(x) = \texttt{orientation}_{\texttt{x}}(x) = \begin{bmatrix} 0& 0& -1\end{bmatrix}^T$. Then two alignment constraints $c_2(x) = \texttt{z\_position}(x) = z_{\text{desired}}$ and $c_3(x) = \texttt{y\_position}(x) \pm \texttt{x\_position}(x) + \texttt{offset}= 0$ enforce the end of the sticks to properly align with the grooves on the block. Each trial is randomized by sampling the initial and goal locations of the sticks within a neighborhood around the grooves’ endpoints to encourage variation in the setup while keeping the problem challenging. Figure \ref{fig:doorway} illustrates an example of the setup.

Table \ref{tab:doorway} records the results of
the doorway-crossing experiment.  Without an effective safety constraint or a trajectory planner, \textbf{\Opt w/o HJR} and \textbf{AtlasRRT (decentralized)} frequently collide. The presence of several manifold constraints also causes \textbf{AtlasRRT (centralized)} to constantly fail by hitting the 20-second time limit and only return a partial trajectory that does not lead to the goal. In contrast, \textbf{\methodname{}} outperforms the baselines in achieving success by a large margin for its ability to identify unsafe situations under complex manifold constraints.

\begin{table}[htbp]
    \centering
\begin{adjustbox}{width=0.98\columnwidth,center}
    \begin{tabular}{ccccc}
    \toprule
    Methods & SR\% $\uparrow$ & CR\% $\downarrow$  & Time [s] $\downarrow$ & PL [rad] $\downarrow$ \\\midrule  
    \methodname{} (ours) &  \textbf{71} & 1 & 0.14 ± 0.14 & 4.7 ± 1.3  \\ 
    \Opt w/o HJR & 9 & 91 & \textbf{0.072 ± 0.011} & \textbf{4.5 ± 0.49} \\ 
    AtlasRRT (decentralized) & 8 & 92 & 1.7 ± 1.9  & 4.7 ± 0.46 \\
    AtlasRRT (centralized) & 0 & \textbf{0} & 22 ± 0.72 & - \\
    % AtlasRRT (joint, 100) & 0 & 25 & 100 ± 5.1 & - \\
    \bottomrule
    \end{tabular}

\end{adjustbox}
    \caption{The success rate, collision rate,  planning time, and mean path length of each method on the \textbf{doorway-crossing} task. Best results for each metric are shown in bold.  }
    \label{tab:doorway}
\end{table}

\section{Conclusion and Discussion}

This paper introduces \methodname{}, a framework that solves constrained HJR to enable decentralized manifold-constrained multi-agent motion planning. The resulting value function captures safety conditions under manifold constraints and enables a receding-horizon trajectory optimization scheme that avoids collisions while respecting task-induced equality constraints. We demonstrate the effectiveness of our method on challenging multi-agent motion planning problems involving diverse task constraints.

Several limitations remain and point to future directions. First, discretized trajectories may drift off the constraint manifold due to solver approximations or inaccurate dynamics, and trajectory segments between planning updates are not explicitly constrained to the manifold. Addressing this issue may require integrating a high-frequency low-level controller with the planner. Second, while the adversarial formulation of HJR removes assumptions on other agents’ control policies for MAMP, it also introduces conservatism, as reflected in the longer paths observed in our experiments. Under the velocity-control setting considered here, applying the fail-safe control inferred from the HJR formulation can also lead to jittery motions. Finally, several sources of uncertainty may affect real-world deployment. For example, the use of neural networks sacrifices formal safety guarantees, and solution quality cannot be directly verified due to the lack of ground-truth solutions in high-dimensional HJR problems. In addition, perception noise, delayed or inaccurate state estimates, and dynamics mismatch may introduce further challenges. Our empirical validation is also limited to velocity-controlled agents and relatively structured scenarios; evaluating the approach on more complex robotic systems remains an important direction. Future work aims to address these challenges, for example by incorporating calibration and uncertainty quantification methods \cite{pmlr-v242-lin24a, lin2025robust}.

\bibliographystyle{IEEEtran}
\bibliography{references}

\end{document}